\documentclass[11pt,a4paper]{article}
\usepackage[hyperref]{emnlp2020}

\usepackage{times}
\usepackage{latexsym}

\usepackage{amsmath,amsfonts,bm}

\def\eqref#1{equation~\ref{#1}}

\def\1{\bm{1}}

\def\eps{{\epsilon}}

\DeclareMathAlphabet{\mathsfit}{\encodingdefault}{\sfdefault}{m}{sl}
\SetMathAlphabet{\mathsfit}{bold}{\encodingdefault}{\sfdefault}{bx}{n}

\usepackage{amsthm}

\usepackage{tabularx} 
\usepackage{amssymb}
\usepackage{pifont}

\usepackage{ragged2e}
\usepackage{booktabs}

\newcommand{\cmark}{\ding{51}}%
\newcommand{\xmark}{\ding{55}}%

\usepackage{hyperref}
\usepackage{url}
\usepackage{dirtytalk}

\usepackage{graphicx} 
\usepackage{float} 
\usepackage{subfigure} 
\usepackage{amsmath}
\usepackage{color,xcolor}
\usepackage{array}
\usepackage{xspace}
\usepackage{multirow}
\usepackage{booktabs}

\usepackage{url}
\usepackage{wrapfig}
\usepackage{enumitem}

\newcommand{\cref}[1]{Condition~(\ref{#1})}

\include{macros}
\def\x{{\mathbf x}}

\usepackage{placeins}

\usepackage{lipsum}
\usepackage{titlesec}

\titlespacing\section{0pt}{7pt plus 0pt minus 1pt}{1pt plus 0pt minus 1pt}
\titlespacing\subsection{0pt}{7pt plus 0pt minus 1pt}{0pt plus 0pt minus 0pt}
\titlespacing\subsubsection{0pt}{4pt plus 0pt minus 1pt}{0pt plus 0pt minus 0pt}
\titlespacing{\paragraph}{0pt}{1pt}{2pt}[0pt]  
\usepackage{etoolbox}
\makeatletter
\preto{\@tabular}{\parskip=2pt}
\makeatother

\usepackage{enumitem}
\setlist[itemize]{leftmargin=*}
\setlist{nosep}

\usepackage[utf8]{inputenc} 

\allowdisplaybreaks

\usepackage{caption}

\usepackage{amsmath,amssymb,amsfonts}
\usepackage{textcomp}
\usepackage{url}  
\usepackage{graphicx}  
\usepackage{multirow}
\usepackage{array}
\usepackage{placeins}

\graphicspath{{../figs/}{../figures/}{../Resources/}{Resources/}}

\usepackage{hhline}

\usepackage{flushend}

\newcommand{\code}{\texttt}

\usepackage{threeparttable}

\newcommand{\TfName}{\textsc{TextFooler}}
\newcommand{\TfAdjName}{\textsc{TFAdjusted}}
\newcommand{\AlzName}{\textsc{GeneticAttack}}
\newcommand{\LibName}{TextAttack}

\newcommand{\QTODO}[1]{}
\newcommand{\ETODO}[1]{}
\newcommand{\ETODISC}[1]{}
\newcommand{\JTODO}[1]{}
\newcommand{\yfnote}[1]{}
\newcommand{\jlnote}[1]{}
\newcommand{\CAMTODO}[1]{}

\renewcommand{\cite}{\citep}

\makeatletter
\newcommand{\removelatexerror}{\let\@latex@error\@gobble}
\makeatother

\usepackage{dblfloatfix}

\usepackage{microtype}

\aclfinalcopy 

\title{Reevaluating Adversarial Examples in Natural Language}

\author{John X. Morris\thanks{\textsuperscript{*} Equal contribution},~ Eli Lifland\footnotemark[1], ~Jack Lanchantin, Yangfeng Ji, Yanjun Qi \\
Department of Computer Science, University of Virginia\\
\texttt{\{jm8wx, edl9cy, jjl5sw, yj3fs, yq2h\}@virginia.edu} \\
}

\date{}

\begin{document}
\maketitle

\begin{abstract}
State-of-the-art attacks on NLP models lack a shared definition of what constitutes a successful attack. These differences make the attacks difficult to compare and hindered the use of adversarial examples to understand and improve NLP models. We distill ideas from past work into a unified framework: a successful natural language adversarial example is a perturbation that fools the model and follows four proposed linguistic constraints. We categorize previous
attacks based on these constraints. For each constraint, we suggest options for
human and automatic evaluation methods. We use these methods to evaluate two
state-of-the-art synonym substitution attacks. We find that perturbations often do not preserve semantics, and 38\% introduce grammatical errors. Next, we
conduct human studies to find a threshold for each evaluation method that
aligns with human judgment. Human surveys reveal that to successfully preserve semantics, we need to significantly increase the minimum cosine similarities between the embeddings of swapped words and between the sentence encodings of original and perturbed sentences. With constraints adjusted to better preserve semantics and grammaticality, the attack success rate drops by over 70 percentage points.
\CAMTODO{Make this abstract less BORING! The introduction is somewhat exciting, but this really isn't.}
\footnote{Our code and datasets are available \href{https://github.com/QData/Reevaluating-NLP-Adversarial-Examples}{here}.}\end{abstract}

\section{Introduction}

\CAMTODO{Look into highlighting text nicely instead of changing its color in some of the tables.}

\CAMTODO{Cite "Elephant in the room" paper as contemporaneous work.}

One way to evaluate the robustness of a machine learning model is to search for inputs that produce incorrect outputs. Inputs intentionally designed to fool deep learning models are referred to as adversarial examples \cite{Goodfellow17Attacking}. Adversarial examples have successfully tricked deep neural networks for image classification: two images that look exactly the same to a human receive completely different predictions from the classifier \cite{goodfellow2014explaining}.

While applicable in the image case, the idea of an indistinguishable change lacks a clear analog in text. Unlike images, two different sequences of text are never entirely indistinguishable. This raises the question: if indistinguishable perturbations are not possible, what are adversarial examples in text?

The literature contains many potential answers to this question, proposing varying definitions for successful adversarial examples \cite{Zhang19Survey}. Even attacks with similar definitions of success often measure it in different ways. The lack of a consistent definition and standardized evaluation has hindered the use of adversarial examples to understand and improve NLP models.  \footnote{We use ‘adversarial example generation methods’ and ‘adversarial attacks’ interchangeably in this  paper.}

Therefore, we propose a unified definition for successful adversarial examples in natural language: perturbations that both fool the model and fulfill a set of linguistic constraints. In Section \ref{sec:constraintsAE}, we present four categories of constraints NLP adversarial examples may follow, depending on the context: semantics, grammaticality, overlap, and non-suspicion to human readers. 

\begin{figure}[t]
    \centering
    \includegraphics[width=.5\textwidth]{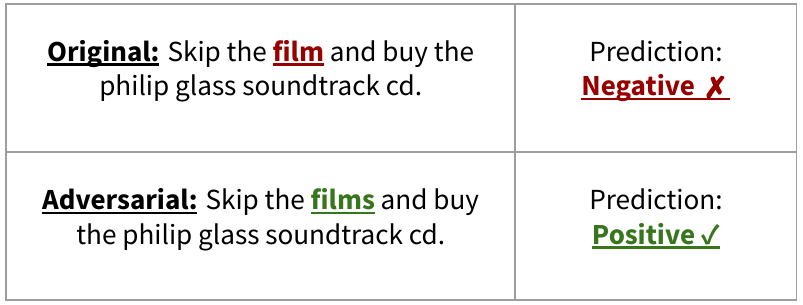}
    \caption{An adversarial example generated by \TfAdjName{} for BERT fine-tuned on the Rotten Tomatoes sentiment analysis dataset. Swapping a single word causes the prediction to change from positive to negative.}
    \label{fig:adv_retraining_acc}
\end{figure}

By explicitly laying out categories of constraints adversarial examples may follow, we introduce a shared vocabulary for discussing constraints on adversarial attacks. In Section~\ref{sec:eval-methods}, we suggest options for human and automatic evaluation methods for each category. We use these methods to evaluate two SOTA synonym substitution attacks: \AlzName{} by \citet{alzantot2018generating} and \TfName{} by \citet{BERT19}. Human surveys show that the perturbed examples often fail to fulfill semantics and non-suspicion constraints. Additionally, a grammar checker detects 39\% more errors in the perturbed examples than in the original inputs, including many types of errors humans almost never make.

In Section~\ref{s5:better:constraints}, we produce \TfAdjName{}, an attack with the same search process as \TfName{}, but with constraint enforcement tuned to generate higher quality adversarial examples. To enforce semantic preservation, we tighten the thresholds on the cosine similarity between embeddings of swapped words and between the sentence encodings of original and perturbed sentences. To enforce  grammaticality, we validate perturbations with a grammar checker. As in \TfName{}, these constraints are applied at each step of the search. Human evaluation shows that \TfAdjName{} generates perturbations that better preserve semantics and are less noticeable to human judges. However, with stricter constraints, the attack success rate decreases from over 80\% to under 20\%. When used for adversarial training, \TfName{}'s examples decreased model accuracy, but \TfAdjName{}'s examples did not.

Without a shared vocabulary for discussing constraints, past work has compared the success rate of search methods with differing constraint application techniques. \citet{BERT19} reported a higher attack success rate for \TfName{} than \citet{alzantot2018generating} did for \AlzName{}, but it was not clear whether the improvement was due to a better search method\footnote{\TfName{} uses a greedy search method with word importance ranking. \AlzName{} uses a genetic algorithm.} or more lenient constraint application\footnote{For example, \TfName{} applies a minimum cosine distance of .5 between embeddings of swapped words. \AlzName{} uses a threshold of .75.}. In Section~\ref{sec:comparing-search} we compare the search methods with constraint application held constant. We find that \AlzName{}'s search method is more successful than \TfName{}'s, contrary to the implications of \citet{BERT19}.

The five main contributions of this paper are:
\begin{itemize}
    \item A definition for constraints on adversarial perturbations in natural language and suggest evaluation methods for each constraint.
    
    \item Constraint evaluations of two SOTA synonym-substitution attacks, revealing that their perturbations often do not preserve semantics, grammaticality, or non-suspicion.
    
    \item Evidence that by aligning automatic constraint application with human judgment, it is possible for attacks to produce successful, valid adversarial examples.
    
    \item Demonstration that reported differences in attack success between \TfName{} and \AlzName{} are the result of more lenient constraint enforcement.
    
    \item Our framework enables fair comparison between attacks, by separating effects of search methods from effects of loosened constraints. 
    
\end{itemize}

\begin{table*}
\centering
\begin{threeparttable}
\centering
\scalebox{0.893}{\begin{tabular}{@{}
>{\raggedleft\arraybackslash}p{2.8cm}|
>{\raggedright\arraybackslash}p{7.1cm}|
>{\raggedright\arraybackslash}p{6.3cm}
}
\toprule
\multicolumn{3}{c}{{\textbf{Input, $\x$}: {"Shall I compare thee to a summer's day?"} – William Shakespeare, Sonnet XVIII}}  \\
\midrule \midrule
\textbf{Constraint} & \centering{\textbf{Perturbation, $\x_{adv}$}} & \multicolumn{1}{c}{\textbf{Explanation}} \\
\hline
\textbf{Semantics} & Shall I compare thee to a {\color{red}winter's} day? & $\x_{adv}$ has a different meaning than $\x$. \\ \hline
\textbf{Grammaticality} & Shall I {\color{red}compares} thee to a summer's day? & $\x_{adv}$ is less grammatically correct than $\x$. \\ \hline
\textbf{Edit Distance} &Sha\color{red}{1}l \color{red}{i} co{\color{red}npp\$haaa}\color{black}{re thee to a} {\color{red}5}umm{\color{red}3}r's day? & $\x$ and $\x_{adv}$ have a large edit distance. \\ \hline
\textbf{Non-suspicion} & \textcolor{red}{Am I gonna} compare thee to a summer's day? & A human reader may suspect this sentence to have been modified. \tnote{1} \\ 
\bottomrule
\end{tabular}}
\scriptsize{
\begin{tablenotes}
\item[1] Shakespeare never used the word ``gonna''. Its first recorded usage wasn't until 1806, and it didn't become popular until the 20th century. 
\end{tablenotes}}
\end{threeparttable}
\caption{\textbf{Adversarial Constraints and Violations.} For each of the four proposed constraints, we show an example for which violates the specified constraint.}
\label{tab:taxonomy-table}
\end{table*}

\section{Constraints on Adversarial Examples in Natural Language}
\label{sec:constraintsAE}


We define $F:\mathcal{X}\rightarrow \mathcal{Y}$ as a predictive model, for example, a deep neural network classifier. $\mathcal{X}$ is the input space and $\mathcal{Y}$ is the output space.
We focus on adversarial perturbations which perturb a correctly predicted input, $\x \in \mathcal{X}$, into an input  $\x_{adv}$. The boolean goal function $G(F, \x_{adv})$ represents whether the goal of the attack has been met. We define ${C_1...C_n}$ as a set of boolean functions indicating whether the perturbation satisfies a certain constraint.

Adversarial attacks search for a perturbation from $\x$ to $\x_{adv}$ which fools $F$ by both achieving some goal, as represented by $G(F,\x_{adv})$, and fulfilling each constraint $C_i(\x,\x_{adv})$.

The definition of the goal function $G$ depends on the purpose of the attack. Attacks on classification frequently aim to either induce any incorrect classification (untargeted) or induce a particular classification (targeted). Attacks on other types of models may have more sophisticated goals. For example, attacks on translation may attempt to change every word of a translation, or introduce targeted keywords into the translation \cite{cheng2018seq2sick}. 

In addition to defining the goal of the attack, the attacker must decide the constraints perturbations must meet. Different use cases require different constraints. We build on the categorization of attack spaces introduced by \citet{Gilmer18Motivating} to introduce a categorization of constraints for adversarial examples in natural language.

In the following, we define four categories of constraints on adversarial perturbations in natural language: semantics, grammatically, overlap, and non-suspicion. Table \ref{tab:taxonomy-table} provides examples of adversarial perturbations that violate each constraint.

\subsection{Semantics}
\label{sec:semantics-constraint}

Semantics constraints require the semantics of the input to be preserved between $\x$ and $\x_{adv}$. Many attacks include constraints on semantics as a way to ensure the correct output is preserved \cite{Zhang19Survey}. As long as the semantics of an input do not change, the correct output will stay the same. There are exceptions: one could imagine tasks for which preserving semantics does not necessarily preserve the correct output. For example, consider the task of classifying passages as written in either Modern or Early Modern English. Perturbing ``why" to ``wherefore" may retain the semantics of the passage, but change the correct label from Modern to Early Modern English\footnote{Wherefore is a synonym for why, but was \href{https://books.google.com/ngrams/graph?content=wherefore&year_start=1500&year_end=2008&corpus=15&smoothing=10&share=&direct_url=t1\%3B\%2Cwherefore\%3B\%2Cc0}{used much more often} centuries ago.}

\subsection{Grammaticality}
\label{sec:grammatical-constraint}

Grammaticality constraints place restrictions on the grammaticality of $\x_{adv}$. For example, an adversary attempting to generate a plagiarised paper which fools a plagiarism checker would need to ensure that the paper remains grammatically correct. Grammatical errors don't necessarily change semantics, as illustrated in Table \ref{tab:taxonomy-table}.

\subsection{Overlap}
\label{sec:overlap-constraint}

Overlap constraints restrict the similarity between $\x$ and $\x_{adv}$ at the character level. This includes constraints like Levenshtein distance as well as n-gram based measures such as BLEU, METEOR and chRF \cite{papineni-etal-2002-bleu,denkowski-lavie-2014-meteor,popovic-2015-chrf}.

Setting a maximum edit distance is useful when the attacker is willing to introduce misspellings. Additionally, the edit distance constraint is sometimes used when improving the robustness of models. For example, \citet{Huang2019AchievingVR} uses Interval Bound Propagation to ensure model robustness to perturbations within some edit distance of the input.

\subsection{Non-suspicion}
\label{sec:non-suspicious-constraint}

Non-suspicion constraints specify that $\x_{adv}$ must appear to be unmodified. Consider the example in Table \ref{tab:taxonomy-table}. While the perturbation preserves semantics and grammar, it switches between Modern and Early Modern English and thus may seem suspicious to readers. 

Note that the definition of the non-suspicious constraint is context-dependent. A sentence that is non-suspicious in the context of a kindergartner's homework assignment might be suspicious in the context of an academic paper. An attack scenario where non-suspicion constraints do not apply is illegal PDF distribution, similar to a case discussed by \citet{Gilmer18Motivating}. Consumers of an illegal PDF may tacitly collude with the person uploading it. They know the document has been altered, but do not care as long as semantics are preserved.
\section{Review and Categorization of SOTA:}
\label{sec:categorization-current}

\paragraph{Attacks by Paraphrase:}\ 
Some studies have generated adversarial examples through paraphrase. \citet{Iyyer18Paraphrase} used neural machine translation systems to generate paraphrases. \citet{ribeiro2018semantically} proposed semantically-equivalent adversarial rules. By definition, paraphrases preserve semantics. Since the systems aim to generate perfect paraphrases, they implicitly follow constraints of grammaticality and non-suspicion.

\paragraph{Attacks by Synonym Substitution:}\ 
Some works focus on an easier way to generate a subset of paraphrases: replacing words from the input with synonyms \cite{alzantot2018generating,BERT19,Kuleshov2018AdversarialEF,papernot2016crafting,ren-etal-2019-generating}. Each attack applies a search algorithm to determine which words to replace with which synonyms. Like the general paraphrase case, they aim to create examples that preserve semantics, grammaticality, and non-suspicion. While not all have an explicit edit distance constraint, some limit the number of words perturbed.

\paragraph{Attacks by Character Substitution:}\ 
Some studies have proposed to attack natural language classification models by deliberately misspelling words \cite{ebrahimi2017hotflip, gao2018black, li2018textbugger}. These attacks use character replacements to change a word into one that the model doesn't recognize. The replacements are designed to create character sequences that a human reader would easily correct into the original words. If there aren't many misspellings, non-suspicion may be preserved. Semantics are preserved as long as human readers can correct the misspellings.

\paragraph{Attacks by Word Insertion or Removal:}\ 
\citet{liang2017deep} and \citet{samanta2017towards} devised a way to determine the most important words in the input and then used heuristics to generate perturbed inputs by adding or removing important words. In some cases, these strategies are combined with synonym substitution. These attacks aim to follow all constraints.

Using constraints defined in Section~\ref{sec:constraintsAE} we categorize a sample of current attacks in Table \ref{tab:categorization-table}.

\begin{table*}[th]
  \centering
\scalebox{0.81}{\begin{tabular}{@{}
>{\raggedleft\arraybackslash}p{9.1cm}
>{\centering\arraybackslash}p{2.1cm}
>{\centering\arraybackslash}p{2.5cm}
>{\centering\arraybackslash}p{2.2cm}
>{\centering\arraybackslash}p{1.8cm}}
\toprule 
    \textbf{Selected Attacks Generating Adversarial Examples in Natural Language} & 
    \textbf{Semantics} & 
    \textbf{Grammaticality}  &
    \textbf{Edit Distance} &
    \textbf{Non-Suspicion}  \\
\midrule \midrule
\textbf{Synonym Substitution.}  \cite{alzantot2018generating,Kuleshov2018AdversarialEF,BERT19,ren-etal-2019-generating} & \cmark  & \cmark & \cmark & \cmark  \\
\hline
\textbf{Character Substitution.} 
\cite{ebrahimi2017hotflip, gao2018black, li2018textbugger} & \cmark  & \ding{53} & \cmark & \cmark \\
\hline
\textbf{Word Insertion or Removal.} 
\cite{liang2017deep, samanta2017towards}  & \cmark  & \cmark & \cmark & \cmark \\
\hline
\textbf{General Paraphrase.} 
\cite{zhao2017generating, ribeiro2018semantically, Iyyer18Paraphrase} & \cmark  & \cmark & \ding{53} & \cmark \\
\hline
\end{tabular}}
\caption{\textbf{Summary of Constraints and Attacks.} This table shows a selection of prior work (rows) categorized by constraints (columns). A ``\cmark'' indicates that the respective attack is supposed to meet the constraint, and a ``\ding{53}''  means the attack is not supposed to meet the constraint.
 \label{tab:categorization-table}}
\end{table*} 

\section{Constraint Evaluation Methods and Case Study}
\label{sec:eval-methods}

\CAMTODO{Can we do label preservation studies?}

For each category of constraints introduced in Section \ref{sec:constraintsAE}, we discuss best practices for both human and automatic evaluation. We leave out overlap due to ease of automatic evaluation. 

Additionally, we perform a case study, evaluating how well black-box synonym substitution attacks \AlzName{} and \TfName{} fulfill constraints. Both attacks find adversarial examples by swapping out words for their synonyms until the classifier is fooled. \AlzName{} uses a genetic algorithm to attack an LSTM trained on the IMDB\footnote{https://datasets.imdbws.com/} document-level sentiment classification dataset. \TfName{} uses a greedy approach to attack an LSTM, CNN, and BERT trained on five classification datasets. We chose these attacks because:
\begin{itemize}
    \item They claim to create perturbations that preserve semantics, maintain grammaticality, and are not suspicious to readers. However, our inspection of the perturbations revealed that many violated these constraints.
    \item They report high attack success rates.\footnote{We use ``attack success rate'' to mean the percentage of the time that an attack can find a successful adversarial example by perturbing a given input. ``After-attack accuracy'' or ``accuracy after attack'' is the accuracy the model achieves after all successful perturbations have been applied.}
    \item They successfully attack two of the most effective models for text classification: LSTM and BERT.
\end{itemize}

To generate examples for evaluation, we attacked BERT using \TfName{} and attacked an LSTM using \AlzName{}. We evaluate both methods on the IMDB dataset. In addition, we evaluate \TfName{} on the Yelp polarity document-level sentiment classification dataset and the Movie Review (MR) sentence-level sentiment classification dataset \cite{pang-lee-2005-seeing,zhang2015character}. We use $1,000$ examples from each dataset. Table \ref{tab:case-study-table} shows example violations of each constraint.
 
\begin{table*}
\centering
\begin{threeparttable}
\centering
\scalebox{0.87}{\begin{tabular}{@{}
>{\raggedleft\arraybackslash}p{3.5cm}|
>{\raggedright\arraybackslash}p{6cm}|
>{\raggedright\arraybackslash}p{6cm}
}
\toprule
\textbf{Constraint Violated} & \centering{\textbf{Input, $\x$}} &\multicolumn{1}{c}{\textbf{Perturbation, $\x_{adv}$}} \\
\hline \hline
\textbf{Semantics} & Jagger, Stoppard and director Michael Apted deliver a \textcolor{blue}{riveting} and surprisingly \textcolor{blue}{romantic ride}. & Jagger, Stoppard and director Michael Apted deliver a \textcolor{red}{baffling} and surprisingly \textcolor{red}{sappy motorbike}.\\ \hline
\textbf{Grammaticality} & A \textcolor{blue}{grating, emaciated} flick. &  A \textcolor{red}{grates, lanky} flick.  \\ \hline
\textbf{Non-suspicion} & \textcolor{blue}{Great} character interaction. & \textcolor{red}{Gargantuan} character interaction. \\ 
\bottomrule
\end{tabular}}
\end{threeparttable}
\caption{\textbf{Real World Constraint Violation Examples.} Perturbations by \TfName{} against BERT fine-tuned on the MR dataset. Each $\x$ is classified as positive, and each $\x_{adv}$ is classified as negative. \label{tab:case-study-table}}
\end{table*}

\subsection{Evaluation of Semantics}
\label{sec:eval_semantics}

\subsubsection{Human Evaluation}
\label{sec:eval_semantics_human_s4}

A few past studies of attacks have included human evaluation of semantic preservation \cite{ribeiro2018semantically,Iyyer18Paraphrase,alzantot2018generating,BERT19}. However, studies often simply ask users to simply rate the ``similarity'' of $\x$ and $\x_{adv}$. We believe this phrasing does not generate an accurate measure of semantic preservation, as users may consider two sentences with different semantics ``similar'' if they only differ by a few words. Instead, users should be explicitly asked whether changes between $\x$ and $\x_{adv}$ preserve the meaning of the original passage.

We propose to ask human judges to rate if meaning is preserved on a Likert scale of 1-5, where 1 is ``Strongly Disagree'' and 5 is ``Strongly Agree'' \cite{likert1932technique}. A perturbation is semantics-preserving if the average score is at least $\epsilon_{sem}$. We propose $\epsilon_{sem} = 4$ as a general rule: on average, humans should at least ``Agree'' that $\x$ and $\x_{adv}$ have the same meaning.

\subsubsection{Automatic Evaluation}

Automatic evaluation of semantic similarity is a well-studied NLP task. The STS Benchmark is used as a common measurement \cite{cer-etal-2017-semeval}. 
 
\citet{Michel19Eval} explored the use of common evaluation metrics for machine translation as a proxy for semantic similarity in the attack setting. While n-gram overlap based approaches are computationally cheap and  work well in the machine translation setting, they do not correlate with human judgment as well as sentence encoders \cite{wieting2018paranmt}. 
 
Some attacks have used sentence encoders to encode two sentences into a pair of fixed-length vectors, then used the cosine distance between the vectors as a proxy for semantic similarity. \TfName{} uses the Universal Sentence Encoder (USE), which achieved a Pearson correlation score of $0.782$ on the STS benchmark \cite{Cer2018UniversalSE}. Another option is BERT fine-tuned for semantic similarity, which achieved a score of $0.865$ \cite{devlin2018bert}. 

Additionally, synonym substitution methods, including \TfName{} and \AlzName{}, often require that words be substituted only with neighbors in the counter-fitted embedding space, which is designed to push synonyms together and antonyms apart \cite{Mrksic2016CounterfittingWV}. 
These automatic metrics of similarity produce a score that represents the similarity between $\x$ and $\x_{adv}$. Attacks depend on a minimum threshold value for each metric to determine whether the changes between $\x$ and $\x_{adv}$ preserve semantics. Human evaluation is needed to find threshold values such that people generally "agree" that semantics is preserved.

\subsubsection{Case Study}

To quantify semantic similarity of $\x$ and $x_{adv}$, we asked users whether they agreed that the changes between the two passages preserved meaning on a scale of 1 (Strongly Disagree) to 5 (Strongly Agree). We averaged scores for each attack method to determine if the method generally preserves semantics.

Perturbations generated by \TfName{} were rated an average of $\textbf{3.28}$, while perturbations generated by \AlzName{} were rated on average $\textbf{2.70}$.\footnote{We hypothesize that \TfName{} achieved higher scores due to its use of USE.} The average rating given for both methods was significantly less than our proposed $\epsilon_{sem}$ of $4$. Using a clear survey question illustrates that humans, on average, don't assess these perturbations as semantics-preserving.  
\subsection{Evaluation of Grammaticality}

\subsubsection{Human Evaluation}

Both \citet{BERT19} and \citet{Iyyer18Paraphrase} reported a human evaluation of grammaticality, but neither study clearly asked if any errors were introduced by a perturbation. For human evaluation of the grammaticality constraint, we propose presenting $\x$ and $\x_{adv}$ together and asking judges if grammatical errors were introduced by the changes made. However, due to the rule-based nature of grammar, automatic evaluation is preferred.

\subsubsection{Automatic Evaluation}

The simplest way to automatically evaluate grammatical correctness is with a rule-based grammar checker. Free grammar checkers are available online in many languages. One popular checker is LanguageTool, an open-source proofreading tool \cite{languagetool}. LanguageTool ships with thousands of human-curated rules for the English language and provides an interface for identifying grammatical errors in sentences. LanguageTool uses rules to detect grammatical errors, statistics to detect uncommon sequences of words, and language model perplexity to detect commonly confused words.

\subsubsection{Case Study}
We ran each of the generated $(\x, \x_{adv})$ pairs through LanguageTool to count grammatical errors. LanguageTool detected more grammatical errors in $\x_{adv}$ than $\x$ for $\textbf{50\%}$ of perturbations generated by \TfName{}, and $\textbf{32\%}$ of perturbations generated by \AlzName{}.

Additionally, perturbations often contain errors that humans rarely make. LanguageTool detected 6 categories for which errors in the perturbed samples appear at least 10 times more frequently than in the original content. Details regarding these error categories and examples of violations are shown in Table \ref{grammar-errors}.

\begin{table*}[th]
  \centering
\scalebox{0.71}{\begin{tabular}{@{}
>{\raggedright\arraybackslash}p{2.4cm}
>{\centering\arraybackslash}p{.8cm}
>{\centering\arraybackslash}p{.8cm}
>{\raggedright\arraybackslash}p{9cm}
>{\raggedright\arraybackslash}p{8cm}}
\toprule 
\textbf{Grammar Rule ID} & $\x$ & $\x_{adv}$ & \textbf{Explanation} & \textbf{Context} \\
\midrule
        \code{TO\_NON\_BASE} & 2 & 123 & Did you mean ``know"? || Replace  with one of [know] & ...ees at person they don't really want to {\color{red} \textbf{knew}} \\
        \code{PRP\_VBG} & 3 & 112 & Did you mean ``we're wanting", ``we are wanting", or ``we were wanting"? || Replace with one of [we're wanting,we are wanting,we were wanting] & while {\color{red} \textbf{we wanting}} macdowell's character to retrieve her h... \\
        \code{A\_PLURAL} & 20 & 294 & Don't use indefinite articles with plural words. Did you mean ``a grate", ``a gratis" or simply ``grates"? || Replace with one of [a grate,a gratis,grates] & a {\color{red} \textbf{grates}}, lanky flick \\
        \code{DID\_BASEFORM} & 25 & 328 & The verb `can't' requires base form of this verb: ``compare" || Replace with one of [compare] &  ...first two cinema in the series, i can't {\color{red}\textbf{compares}} friday after next to them, but nothing ... \\
        \code{PRP\_VB} & 6 & 73 & Do not use a noun immediately after the pronoun `it'. Use a verb or an adverb, or possibly some other part of speech. || Replace game with one of [] &   ...ble of being gravest, so thick with wry it {\color{red}\textbf{game}} like a readings from bartlett's familia... \\
         \code{PRP\_MD\_NN} &  4 &  46 & It seems that a verb or adverb has been misspelled or is missing here. || Replace with one of [can be appreciative,can have appreciative] & ...y bit as awful as borchardt's coven, we can {\color{red} \textbf{appreciative}} it anyway \\
        \code{NON3PRS\_VERB} & 7 & 78 & The pronoun 'they' must be used with a non-third-person form of a verb: ``do" || Replace with one of [do] & they {\color{red}\textbf{does}} a ok operating of painting this family ... \\
    \bottomrule
    \end{tabular}}
    \caption{\textbf{Adversarial Examples Contain Uncommon Grammatical Errors.} This table shows grammatical errors detected by LanguageTool that appeared far more often in the perturbed samples. $\x$ and $\x_{adv}$ denote the numbers of errors detected in $\x$ and $\x_{adv}$ across 3,115 examples generated by \TfName{} and \AlzName{}.
    \CAMTODO{languagetoolpython incorrectly says "Replace with one of ....". Fix that and then rerun to get the correct replacement words}  \QTODO{this table may go to Appendix.}
    \label{grammar-errors}}
\end{table*}

\subsection{Evaluation of Non-suspicion}
\label{sec:eval_ns}

\subsubsection{Human Evaluation}
We propose evaluation of non-suspicion by having judges view a shuffled mix of real and adversarial inputs and guess whether each is real or computer-altered. This is similar to the human evaluation done by \citet{ren-etal-2019-generating}, but we formulate it as a binary classification task rather than on a 1-5 scale. A perturbed example $\x_{adv}$ is not suspicious if the percentage of judges who identify $\x_{adv}$ as computer-altered is at most $\epsilon_{ns}$, where $0 \leq \epsilon_{ns} \leq 1$.

\subsubsection{Automatic Evaluation}
Automatic evaluation may be used to guess whether or not an adversarial example is suspicious. Models can be trained to classify passages as real or perturbed, just as human judges do. For example, \citet{CoLA-task} trained sentence encoders on a real/fake task as a proxy for evaluation of linguistic acceptability. Recently, \citet{Zellers19NFN} demonstrated that GROVER, a transformer-based text generation model, could classify its own generated news articles as human or machine-written with high accuracy. \CAMTODO{Consider citing GPT-3 here.}

\subsubsection{Case Study}
We presented a shuffled mix of real and perturbed examples to human judges and asked if they were real or computer-altered. As this is a time-consuming task for long documents, we only evaluated adversarial examples generated by \TfName{} on the sentence-level MR dataset.

If all generated examples were non-suspicious, judges would average 50\% accuracy, as they would not be able to distinguish between real and perturbed examples. In this case, judges achieved $\textbf{69.2\%}$ accuracy.

\section{Producing Higher Quality Adversarial Examples} 
\label{s5:better:constraints}

In Section \ref{sec:eval-methods}, we evaluated how well generated examples met constraints. We found that although attacks in NLP aspire to meet linguistic constraints, in practice, they frequently violate them. Now, we adjust automatic constraints applied during the course of the attack to produce better quality adversarial examples.

We set out to find if a set of constraint application methods with appropriate thresholds could produce adversarial examples that are semantics-preserving, grammatical and non-suspicious. We modified \TfName{} to produce \TfAdjName{}, a new attack with stricter constraint application. To enforce grammaticality, we added LanguageTool. To enforce semantic preservation, we tuned two thresholds which filter out invalid word substitutions: (a) minimum cosine similarity between counter-fitted word embeddings and (b) minimum cosine similarity between sentence embeddings. Through human studies, we found threshold values of $\textbf{0.9}$ for (a) and $\textbf{0.98}$ for (b)\footnote{Details in the appendix, Section A.2.2.}. We implemented \TfAdjName{} using \LibName{}, a Python framework for implementing adversarial attacks in NLP \cite{Morris2020TextAttack}.
\subsection{With Adjusted Constraint Application}
We tested \TfAdjName{} to determine the effect of tightening constraint application. We used the IMDB, Yelp, and MR datasets for classifcation as in Section \ref{sec:eval-methods}.  We added the SNLI and MNLI entailment datasets \cite{DBLP:journals/corr/BowmanAPM15,williams-etal-2018-broad} for the portions not requring human evaluation. Table \ref{tab:after-tuning-succ} shows the results.

\noindent \textbf{Semantics.} \TfName{} generates perturbations for which human judges are on average ``Not sure" if semantics are preserved. With perturbations generated by \TfAdjName{}, human judges on average ``Agree" that semantics are preserved.

\noindent \textbf{Grammaticality.} Since all examples produced by \TfAdjName{} are checked with LanguageTool, no perturbation can introduce grammatical errors. \footnote{Since the MR dataset is already lowercased and tokenized, it is difficult for a rule-based grammar checker like LanguageTool to parse some inputs.}

\noindent \textbf{Non-suspicion.} We repeated the non-suspicion study from Section \ref{sec:eval_ns} with the examples generated by \TfAdjName{}. Participants were able to guess with $58.8\%$ accuracy whether inputs were computer-altered. The accuracy is over $10\%$ lower than the accuracy on the examples generated by \TfName{}.

\noindent \textbf{Attack success.} For each of the three datasets, the attack success rate decreased by at least $71$ percentage points (see last row of Table \ref{tab:after-tuning-succ}).

\begin{table*}[tb]
    \centering
\scalebox{0.86}{\begin{tabular}{|p{4.8cm}|ccccc|p{4.8cm}|}
        \hline
        \multicolumn{1}{|r|}{Datasets $\longrightarrow$} & IMDB & Yelp & MR & SNLI & MNLI & Note \\
        \hhline{|=|===|==|=|} 
         Semantic Preservation (before)  & $3.41$ & $3.05$ & $3.37$ & $-$ & $-$ & \\
         Semantic Preservation (after)  & $4.06$ & $3.94$ & $4.18$ & $-$ & $-$ & Higher value: more preserved \\
        \hline
         Grammatical Error \% (before) & $52.8$ & $61.2$ & $28.3$ & $26.7$ & $20.1$ & \\
         Grammatical Error \% (after)   & $0$ & $0$ & $0$ & $0$ & $0$ & Lower value: less mistakes \\
        \hline
         Non-suspicion \% (before) & $-$ & $-$ & $69.2$ & $-$ & $-$ & \\
         Non-suspicion \% (after)  & $-$ & $-$ & $58.8$ & $-$ & $-$ & Lower value: less suspicious\\
        \hline
         Attack Success \% (before) & $85.0$ & $93.2$ & $86.6$ & $94.5$ & $95.1$ & \\
         Attack Success \% (after)  & $13.9$ & $5.3$ & $10.6$ & $7.2$ & $14.8$ &  \\
         Difference (before - after)  & $\bm{71.1}$ & $\bm{87.9}$ & $\bm{76.0}$ & $\bm{87.3}$ & $\bm{80.3}$ & \\
         \hline
    \end{tabular}}
    \caption{Results from running \TfName{} (before) and \TfAdjName{} (after). Attacks are on BERT classification models fine-tuned for five respective NLP datasets.}
    \label{tab:after-tuning-succ}
\end{table*}

\subsection{Adversarial Training With Higher Quality Examples} 

Using the $9,595$ samples in the MR training set as seed inputs, \TfName{} generated \textbf{7,382} adversarial examples, while \TfAdjName{} generated just \textbf{825}. We append each set of adversarial examples to a copy of the original MR training set and fine-tuned a pre-trained BERT model for 10 epochs. Figure \ref{fig:adv_retraining_acc} plots the test accuracy over 10 training epochs, averaged over 5 random seeds per dataset. While neither training method strongly impacts accuracy, the augmentation using \TfAdjName{} has a better impact than that of \TfName{}.

\begin{figure}[t]
    \centering
    \includegraphics[width=.4\textwidth]{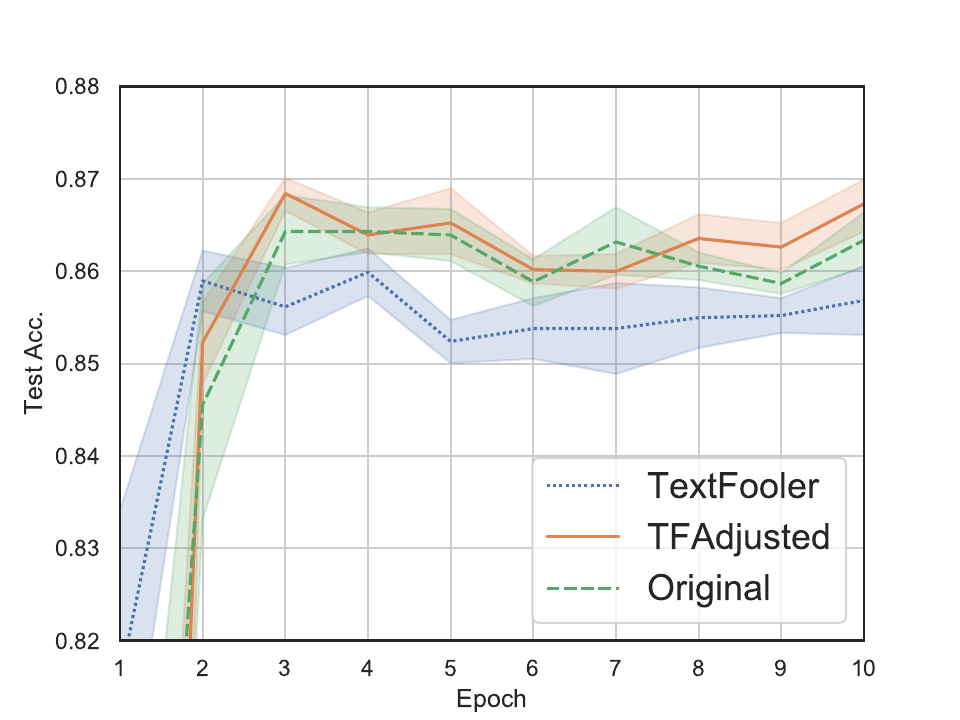}
    \caption{Accuracy of adversarially trained models on the MR test set. Augmentation with adversarial examples generated by \TfName{} (blue), although higher in quantity, decreases the overall test accuracy while examples generated by \TfAdjName{} (orange) have a small positive effect.}
    \CAMTODO{Mention BERT in this caption}
    \label{fig:adv_retraining_acc}
\end{figure}

We then re-ran the two attacks using 1000 examples from the MR test set as seeds. Again averaging over 5 random seeds, we found no significant change in robustness. That is, models trained on the original MR dataset were approximately as robust as those trained on the datasets augmented with \TfName{} and \TfAdjName{} examples. This corroborates the findings of \citet{alzantot2018generating} and contradicts those of \citet{BERT19}. We include further analysis along with some hypotheses for the discrepancies in adversarial training results in A.4. \CAMTODO{Link to appendix!}
\subsection{Ablation of \TfAdjName{} Constraints}
\label{sec:ablation}
\TfAdjName{} generated better quality adversarial examples by constraining its search to exclude examples that fail to meet three constraints: word embedding distance, sentence encoder similarity, and grammaticality. We performed an ablation study to understand the relative impact of each on attack success rate.

\begin{table}
    \centering
\scalebox{0.77}{\begin{tabular}{|p{3.5cm}|ccccc|}
        \hline
        \textbf{Constraint Removed} & \textbf{Yelp} & \textbf{IMDB} & \textbf{MR} & \textbf{MNLI} & \textbf{SNLI} \\
         \hline
        (Original - all used) & 5.3 & 13.9 & 10.6 & 14.3 & 7.2  \\
        \hline
        Sentence Encoding & 22.9 & 45.0 & 28.7 & 44.4 & 31.2 \\
         \hline
        Word Embedding & \textbf{74.6} & \textbf{87.1} & \textbf{52.9} & \textbf{82.7} & \textbf{69.8} \\
        \hline
        Grammar Checking & 5.8 & 15.0 & 11.6 & 15.4 & 9.0  \\
         \hline
    \end{tabular}}
    \caption{Ablation study: effect of removal of a single constraint on \TfAdjName{} attack success rate. Attacks against BERT fine-tuned on each dataset.}
    \label{tab:constraint-ablation-study}
\end{table}

We reran three \TfAdjName{} attacks (one for each constraint removed) on each dataset. Table \ref{tab:constraint-ablation-study} shows attack success rate after individually removing each constraint. The word embedding distance constraint was the greatest inhibitor of attack success rate, followed by the sentence encoder. 

\CAMTODO{Run ablation study moving constraints from TextFooler levels to Tf-Adjusted levels.}

\section{Comparing Search Methods}
\label{sec:comparing-search}

When an attack's success rate improves, it may be the result of either (a) improvement of the search method for finding adversarial perturbations or (b) more lenient constraint definitions or constraint application. \TfName{} achieves a higher success rate than \AlzName{}, but \citet{BERT19} did not identify whether the improvement was due to (a) or (b). Since \TfName{} uses both a different search method and different constraint application methods than \AlzName{}, the source of the difference in attack success rates is unclear.

To determine which search method is more effective, we used \LibName{} to compose attacks from the search method of \AlzName{} and the constraint application methods of each of \TfName{} and \TfAdjName{} \cite{Morris2020TextAttack}. With the constraint application held constant, we can identify the source of the difference in attack success rate. Table \ref{tab:attack-relative-tuning} reveals that the genetic algorithm of \AlzName{} is more successful than the greedy search of \TfName{} at both constraint application levels. This reveals the source of improvement in attack success rate between \AlzName{} and \TfName{} to be more lenient constraint application. However, \AlzName{}'s genetic algorithm is far more computationally expensive, requiring over $40$x more model queries.

\begin{table*}[thb]
    \centering
\scalebox{.8}{\begin{tabular}{|c|cc|cc|}
        \hline
        Constraints & \multicolumn{2}{c|}{\TfAdjName{}} & \multicolumn{2}{c|}{\TfName{}} \\
        Search Method &\TfName{} & \AlzName{} & \TfName{} & \AlzName{} \\
         \hhline{|=|==|==|} 
          Semantic Preservation & 4.06 & 4.11 & - & -\\
          Grammatical Error \% & 0 & 0 & - & - \\
          Non-suspicion Score & 58.8 & 56.9 & - & - \\
         \hhline{|=|==|==|} 
         Attack Success \% & 10.6 & \textbf{12.0} & 91.1 & \textbf{95.0}   \\
          Perturbed Word \% & 11.1 & 11.0 & 18.9 & 17.2 \\
         Num Queries & \textbf{27.1} & 4431.6 & \textbf{77.0} & 3225.7  \\
         \hline
    \end{tabular}}
    \caption{Comparison of the search methods from \AlzName{} and \TfName{} with two sets of constraints (\TfName{} and \TfAdjName{}). Attacks were run on 1000 samples against BERT fine-tuned on the MR dataset. \AlzName{}'s genetic algorithm is more successful than \TfName{}'s greedy strategy, albeit much less efficient.}
    \label{tab:attack-relative-tuning}
\end{table*}

\section{Discussion}
\textbf{Tradeoff between attack success and example quality.} \TfAdjName{} made semantic constraints more selective, which helped attacks generate examples that scored above 4 on the Likert scale for preservation of semantics. However, this led to a steep drop in attack success rate. This indicates that, when only allowing adversarial perturbations that preserve semantics and grammaticality, NLP models are relatively robust to current synonym substitution attacks. Note that our set of constraints isn't necessarily optimal for every attack scenario. Some contexts may require fewer constraints or less strict constraint application.

\textbf{Decoupling search methods and constraints.} It is critical that researchers decouple new search methods from new constraint evaluation and constraint application methods. Demonstrating the performance of a new attack that simultaneously introduces a new search method and new constraints makes it unclear whether empirical gains indicate a more effective attack or a more relaxed set of constraints. This mirrors a broader trend in machine learning where researchers report differences that come from changing multiple independent variables, making the sources of empirical gains unclear \cite{Lipton2018TroublingTrends}. This is especially relevant in adversarial NLP, where each experiment depends on many parameters.

\textbf{Towards improved methods for generating textual adversarial examples.} As models improve at paraphrasing inputs, we will be able to explore the space of adversarial examples beyond synonym substitutions. As models improve at measuring semantic similarity, we will be able to more rigorously ensure that adversarial perturbations preserve semantics. It remains to be seen how robust BERT is when subject to paraphrase attacks that rigorously preserve semantics and grammaticality.

\section{Related Work}
The goal of creating adversarial examples that preserve semantics and grammaticality is common in the NLP attack literature \cite{Zhang19Survey}. However, previous works use different definitions of adversarial examples, making it difficult to compare methods. We provide a unified definition of an adversarial example based on a goal function and a set of linguistic constraints.

\citet{Gilmer18Motivating} laid out a set of potential constraints for the attack space when generating adversarial examples, which are each useful in different real-world scenarios. However, they did not discuss NLP attacks in particular. \citet{Michel19Eval} defined a framework for evaluating attacks on machine translation models, focusing on meaning preservation constraints, but restricted their definitions to sequence-to-sequence models. Other research on NLP attacks has suggested various constraints but has not introduced a shared vocabulary and categorization that allows for effective comparisons between attacks.

\section{Conclusion}

We showed that two state-of-the-art synonym substitution attacks, \TfName{} and \AlzName{}, frequently violate the constraints they claim to follow. We created \TfAdjName{}, which applies constraints that produce adversarial examples judged to preserve semantics and grammaticality. 

Due to the lack of a shared vocabulary for discussing NLP attacks, the source of improvement in attack success rate between \TfName{} and \AlzName{} was unclear. Holding constraint application constant revealed that the source of \TfName{}'s improvement was lenient constraint application (rather than a better search method). With a shared framework for defining and applying constraints, future research can focus on developing better search methods and better constraint application techniques for preserving semantics and grammaticality.

\bibliographystyle{acl_natbib}
\bibliography{0-EMNLP-2020}

\begin{thebibliography}{39}
\expandafter\ifx\csname natexlab\endcsname\relax\def\natexlab#1{#1}\fi

\bibitem[{Alzantot et~al.(2018)Alzantot, Sharma, Elgohary, Ho, Srivastava, and
  Chang}]{alzantot2018generating}
Moustafa Alzantot, Yash Sharma, Ahmed Elgohary, Bo-Jhang Ho, Mani Srivastava,
  and Kai-Wei Chang. 2018.
\newblock Generating natural language adversarial examples.
\newblock \emph{arXiv preprint arXiv:1804.07998}.

\bibitem[{Bowman et~al.(2015)Bowman, Angeli, Potts, and
  Manning}]{DBLP:journals/corr/BowmanAPM15}
Samuel~R. Bowman, Gabor Angeli, Christopher Potts, and Christopher~D. Manning.
  2015.
\newblock \href {http://arxiv.org/abs/1508.05326} {A large annotated corpus for
  learning natural language inference}.
\newblock \emph{CoRR}, abs/1508.05326.

\bibitem[{Cer et~al.(2017)Cer, Diab, Agirre, Lopez-Gazpio, and
  Specia}]{cer-etal-2017-semeval}
Daniel Cer, Mona Diab, Eneko Agirre, I{\~n}igo Lopez-Gazpio, and Lucia Specia.
  2017.
\newblock \href {https://doi.org/10.18653/v1/S17-2001} {{S}em{E}val-2017 task
  1: Semantic textual similarity multilingual and crosslingual focused
  evaluation}.
\newblock In \emph{Proceedings of the 11th International Workshop on Semantic
  Evaluation ({S}em{E}val-2017)}, pages 1--14, Vancouver, Canada. Association
  for Computational Linguistics.

\bibitem[{Cer et~al.(2018)Cer, Yang, yi~Kong, Hua, Limtiaco, John, Constant,
  Guajardo-Cespedes, Yuan, Tar, Sung, Strope, and
  Kurzweil}]{Cer2018UniversalSE}
Daniel Cer, Yinfei Yang, Sheng yi~Kong, Nan Hua, Nicole Limtiaco, Rhomni~St.
  John, Noah Constant, Mario Guajardo-Cespedes, Steve Yuan, Chris Tar,
  Yun-Hsuan Sung, Brian Strope, and Ray Kurzweil. 2018.
\newblock Universal sentence encoder.
\newblock \emph{ArXiv}, abs/1803.11175.

\bibitem[{Cheng et~al.(2018)Cheng, Yi, Zhang, Chen, and
  Hsieh}]{cheng2018seq2sick}
Minhao Cheng, Jinfeng Yi, Huan Zhang, Pin-Yu Chen, and Cho-Jui Hsieh. 2018.
\newblock Seq2sick: Evaluating the robustness of sequence-to-sequence models
  with adversarial examples.
\newblock \emph{arXiv preprint arXiv:1803.01128}.

\bibitem[{Denkowski and Lavie(2014)}]{denkowski-lavie-2014-meteor}
Michael Denkowski and Alon Lavie. 2014.
\newblock \href {https://doi.org/10.3115/v1/W14-3348} {Meteor universal:
  Language specific translation evaluation for any target language}.
\newblock In \emph{Proceedings of the Ninth Workshop on Statistical Machine
  Translation}, pages 376--380, Baltimore, Maryland, USA. Association for
  Computational Linguistics.

\bibitem[{Devlin et~al.(2018)Devlin, Chang, Lee, and
  Toutanova}]{devlin2018bert}
Jacob Devlin, Ming-Wei Chang, Kenton Lee, and Kristina Toutanova. 2018.
\newblock Bert: Pre-training of deep bidirectional transformers for language
  understanding.
\newblock \emph{arXiv preprint arXiv:1810.04805}.

\bibitem[{Ebrahimi et~al.(2017)Ebrahimi, Rao, Lowd, and
  Dou}]{ebrahimi2017hotflip}
Javid Ebrahimi, Anyi Rao, Daniel Lowd, and Dejing Dou. 2017.
\newblock Hotflip: White-box adversarial examples for text classification.
\newblock \emph{arXiv preprint arXiv:1712.06751}.

\bibitem[{Gao et~al.(2018)Gao, Lanchantin, Soffa, and Qi}]{gao2018black}
Ji~Gao, Jack Lanchantin, Mary~Lou Soffa, and Yanjun Qi. 2018.
\newblock Black-box generation of adversarial text sequences to evade deep
  learning classifiers.
\newblock In \emph{IEEE Security and Privacy Workshops (SPW)}.

\bibitem[{Gilmer et~al.(2018)Gilmer, Adams, Goodfellow, Andersen, and
  Dahl}]{Gilmer18Motivating}
Justin Gilmer, Ryan~P. Adams, Ian~J. Goodfellow, David Andersen, and George~E.
  Dahl. 2018.
\newblock \href {http://arxiv.org/abs/1807.06732} {Motivating the rules of the
  game for adversarial example research}.
\newblock \emph{CoRR}, abs/1807.06732.

\bibitem[{Goodfellow et~al.(2017)Goodfellow, Papernot, Huang, Duan, Abbeel, and
  Clark}]{Goodfellow17Attacking}
Ian Goodfellow, Nicolas Papernot, Sandy Huang, Rocky Duan, Pieter Abbeel, and
  Jack Clark.
\newblock \href {https://openai.com/blog/adversarial-example-research/}
  {Attacking machine learning with adversarial examples} [online]. 2017.

\bibitem[{Goodfellow et~al.(2014)Goodfellow, Shlens, and
  Szegedy}]{goodfellow2014explaining}
Ian~J Goodfellow, Jonathon Shlens, and Christian Szegedy. 2014.
\newblock Explaining and harnessing adversarial examples.
\newblock \emph{arXiv preprint arXiv:1412.6572}.

\bibitem[{Huang et~al.(2019)Huang, Stanforth, Welbl, Dyer, Yogatama, Gowal,
  Dvijotham, and Kohli}]{Huang2019AchievingVR}
Po-Sen Huang, Robert Stanforth, Johannes Welbl, Chris Dyer, Dani Yogatama, Sven
  Gowal, Krishnamurthy Dvijotham, and Pushmeet Kohli. 2019.
\newblock Achieving verified robustness to symbol substitutions via interval
  bound propagation.
\newblock \emph{ArXiv}, abs/1909.01492.

\bibitem[{Iyyer et~al.(2018)Iyyer, Wieting, Gimpel, and
  Zettlemoyer}]{Iyyer18Paraphrase}
Mohit Iyyer, John Wieting, Kevin Gimpel, and Luke Zettlemoyer. 2018.
\newblock \href {http://arxiv.org/abs/1804.06059} {Adversarial example
  generation with syntactically controlled paraphrase networks}.
\newblock \emph{CoRR}, abs/1804.06059.

\bibitem[{{Jin} et~al.(2019){Jin}, {Jin}, {Tianyi Zhou}, and
  {Szolovits}}]{BERT19}
Di~{Jin}, Zhijing {Jin}, Joey {Tianyi Zhou}, and Peter {Szolovits}. 2019.
\newblock \href {http://arxiv.org/abs/1907.11932} {{Is BERT Really Robust? A
  Strong Baseline for Natural Language Attack on Text Classification and
  Entailment}}.
\newblock \emph{arXiv e-prints}, page arXiv:1907.11932.

\bibitem[{Kuleshov et~al.(2018)Kuleshov, Thakoor, Lau, and
  Ermon}]{Kuleshov2018AdversarialEF}
Volodymyr Kuleshov, Shantanu Thakoor, Tingfung Lau, and Stefano Ermon. 2018.
\newblock \href {https://openreview.net/forum?id=r1QZ3zbAZ} {Adversarial
  examples for natural language classification problems}.

\bibitem[{Li et~al.(2018)Li, Ji, Du, Li, and Wang}]{li2018textbugger}
Jinfeng Li, Shouling Ji, Tianyu Du, Bo~Li, and Ting Wang. 2018.
\newblock Textbugger: Generating adversarial text against real-world
  applications.
\newblock \emph{arXiv preprint arXiv:1812.05271}.

\bibitem[{Liang et~al.(2017)Liang, Li, Su, Bian, Li, and Shi}]{liang2017deep}
Bin Liang, Hongcheng Li, Miaoqiang Su, Pan Bian, Xirong Li, and Wenchang Shi.
  2017.
\newblock Deep text classification can be fooled.
\newblock \emph{arXiv preprint arXiv:1704.08006}.

\bibitem[{Likert(1932)}]{likert1932technique}
R.~Likert. 1932.
\newblock \href {https://books.google.com/books?id=9rotAAAAYAAJ} {\emph{A
  Technique for the Measurement of Attitudes}}.
\newblock Number nos. 136-165 in A Technique for the Measurement of Attitudes.
  publisher not identified.

\bibitem[{Lipton and Steinhardt(2018)}]{Lipton2018TroublingTrends}
Zachary~Chase Lipton and Jacob Steinhardt. 2018.
\newblock Troubling trends in machine learning scholarship.
\newblock \emph{ArXiv}, abs/1807.03341.

\bibitem[{Michel et~al.(2019)Michel, Li, Neubig, and Pino}]{Michel19Eval}
Paul Michel, Xian Li, Graham Neubig, and Juan~Miguel Pino. 2019.
\newblock \href {http://arxiv.org/abs/1903.06620} {On evaluation of adversarial
  perturbations for sequence-to-sequence models}.
\newblock \emph{CoRR}, abs/1903.06620.

\bibitem[{Morris et~al.(2020)Morris, Lifland, Yoo, and
  Qi}]{Morris2020TextAttack}
John~X. Morris, Eli Lifland, Jin~Yong Yoo, and Yanjun Qi. 2020.
\newblock \href {http://arxiv.org/abs/arXiv:2005.05909} {Textattack: A
  framework for adversarial attacks in natural language processing}.

\bibitem[{Mrksic et~al.(2016)Mrksic, S{\'e}aghdha, Thomson, Gasic,
  Rojas-Barahona, hao Su, Vandyke, Wen, and Young}]{Mrksic2016CounterfittingWV}
Nikola Mrksic, Diarmuid~{\'O} S{\'e}aghdha, Blaise Thomson, Milica Gasic,
  Lina~Maria Rojas-Barahona, Pei hao Su, David Vandyke, Tsung-Hsien Wen, and
  Steve~J. Young. 2016.
\newblock Counter-fitting word vectors to linguistic constraints.
\newblock In \emph{HLT-NAACL}.

\bibitem[{Naber(2003)}]{languagetool}
Daniel Naber. 2003.
\newblock A rule-based style and grammar checker.

\bibitem[{Pang and Lee(2005)}]{pang-lee-2005-seeing}
Bo~Pang and Lillian Lee. 2005.
\newblock \href {https://doi.org/10.3115/1219840.1219855} {Seeing stars:
  Exploiting class relationships for sentiment categorization with respect to
  rating scales}.
\newblock In \emph{Proceedings of the 43rd Annual Meeting of the Association
  for Computational Linguistics ({ACL}{'}05)}, pages 115--124, Ann Arbor,
  Michigan. Association for Computational Linguistics.

\bibitem[{Papernot et~al.(2016)Papernot, McDaniel, Swami, and
  Harang}]{papernot2016crafting}
Nicolas Papernot, Patrick McDaniel, Ananthram Swami, and Richard Harang. 2016.
\newblock Crafting adversarial input sequences for recurrent neural networks.
\newblock In \emph{Military Communications Conference, MILCOM 2016-2016 IEEE},
  pages 49--54. IEEE.

\bibitem[{Papineni et~al.(2002)Papineni, Roukos, Ward, and
  Zhu}]{papineni-etal-2002-bleu}
Kishore Papineni, Salim Roukos, Todd Ward, and Wei-Jing Zhu. 2002.
\newblock \href {https://doi.org/10.3115/1073083.1073135} {{B}leu: a method for
  automatic evaluation of machine translation}.
\newblock In \emph{Proceedings of the 40th Annual Meeting of the Association
  for Computational Linguistics}, pages 311--318, Philadelphia, Pennsylvania,
  USA. Association for Computational Linguistics.

\bibitem[{Popovi{\'c}(2015)}]{popovic-2015-chrf}
Maja Popovi{\'c}. 2015.
\newblock \href {https://doi.org/10.18653/v1/W15-3049} {chr{F}: character
  n-gram f-score for automatic {MT} evaluation}.
\newblock In \emph{Proceedings of the Tenth Workshop on Statistical Machine
  Translation}, pages 392--395, Lisbon, Portugal. Association for Computational
  Linguistics.

\bibitem[{Reimers and Gurevych(2019)}]{reimers-2019-sentence-bert}
Nils Reimers and Iryna Gurevych. 2019.
\newblock \href {http://arxiv.org/abs/1908.10084} {Sentence-bert: Sentence
  embeddings using siamese bert-networks}.
\newblock In \emph{Proceedings of the 2019 Conference on Empirical Methods in
  Natural Language Processing}. Association for Computational Linguistics.

\bibitem[{Ren et~al.(2019)Ren, Deng, He, and Che}]{ren-etal-2019-generating}
Shuhuai Ren, Yihe Deng, Kun He, and Wanxiang Che. 2019.
\newblock \href {https://doi.org/10.18653/v1/P19-1103} {Generating natural
  language adversarial examples through probability weighted word saliency}.
\newblock In \emph{Proceedings of the 57th Annual Meeting of the Association
  for Computational Linguistics}, pages 1085--1097, Florence, Italy.
  Association for Computational Linguistics.

\bibitem[{Ribeiro et~al.(2018)Ribeiro, Singh, and
  Guestrin}]{ribeiro2018semantically}
Marco~Tulio Ribeiro, Sameer Singh, and Carlos Guestrin. 2018.
\newblock Semantically equivalent adversarial rules for debugging nlp models.
\newblock In \emph{Proceedings of the 56th Annual Meeting of the Association
  for Computational Linguistics (Volume 1: Long Papers)}, pages 856--865.

\bibitem[{Samanta and Mehta(2017)}]{samanta2017towards}
Suranjana Samanta and Sameep Mehta. 2017.
\newblock Towards crafting text adversarial samples.
\newblock \emph{arXiv preprint arXiv:1707.02812}.

\bibitem[{Warstadt et~al.(2018)Warstadt, Singh, and Bowman}]{CoLA-task}
Alex Warstadt, Amanpreet Singh, and Samuel~R. Bowman. 2018.
\newblock \href {http://arxiv.org/abs/1805.12471} {Neural network acceptability
  judgments}.
\newblock \emph{CoRR}, abs/1805.12471.

\bibitem[{Wieting and Gimpel(2018)}]{wieting2018paranmt}
John Wieting and Kevin Gimpel. 2018.
\newblock Paranmt-50m: Pushing the limits of paraphrastic sentence embeddings
  with millions of machine translations.
\newblock In \emph{Proceedings of the 56th Annual Meeting of the Association
  for Computational Linguistics (Volume 1: Long Papers)}, pages 451--462.

\bibitem[{Williams et~al.(2018)Williams, Nangia, and
  Bowman}]{williams-etal-2018-broad}
Adina Williams, Nikita Nangia, and Samuel Bowman. 2018.
\newblock \href {https://doi.org/10.18653/v1/N18-1101} {A broad-coverage
  challenge corpus for sentence understanding through inference}.
\newblock In \emph{Proceedings of the 2018 Conference of the North {A}merican
  Chapter of the Association for Computational Linguistics: Human Language
  Technologies, Volume 1 (Long Papers)}, pages 1112--1122, New Orleans,
  Louisiana. Association for Computational Linguistics.

\bibitem[{Zellers et~al.(2019)Zellers, Holtzman, Rashkin, Bisk, Farhadi,
  Roesner, and Choi}]{Zellers19NFN}
Rowan Zellers, Ari Holtzman, Hannah Rashkin, Yonatan Bisk, Ali Farhadi,
  Franziska Roesner, and Yejin Choi. 2019.
\newblock \href {http://arxiv.org/abs/1905.12616} {Defending against neural
  fake news}.
\newblock \emph{CoRR}, abs/1905.12616.

\bibitem[{Zhang et~al.(2019)Zhang, Sheng, and Alhazmi}]{Zhang19Survey}
Wei~Emma Zhang, Quan~Z. Sheng, and Ahoud Abdulrahmn~F. Alhazmi. 2019.
\newblock \href {http://arxiv.org/abs/1901.06796} {Generating textual
  adversarial examples for deep learning models: {A} survey}.
\newblock \emph{CoRR}, abs/1901.06796.

\bibitem[{Zhang et~al.(2015)Zhang, Zhao, and LeCun}]{zhang2015character}
Xiang Zhang, Junbo Zhao, and Yann LeCun. 2015.
\newblock Character-level convolutional networks for text classification.
\newblock In \emph{Advances in neural information processing systems}, pages
  649--657.

\bibitem[{Zhao et~al.(2017)Zhao, Dua, and Singh}]{zhao2017generating}
Zhengli Zhao, Dheeru Dua, and Sameer Singh. 2017.
\newblock Generating natural adversarial examples.
\newblock \emph{arXiv preprint arXiv:1710.11342}.

\end{thebibliography}

\cleardoublepage

\appendix
\section{Appendix}
\subsection{Experimental Setup}

We ran our experiments on machines running CentOS 7 with 4 TITAN X Pascal GPUs. We fine-tune models on `BERT-base`, which has 110 million parameters. For all datasets, we used the standard train and test split, with the exception of MR, which comes as a single dataset. As is custom, we split the MR dataset into 90\% training data and 10\% testing data. The samples we chose for each dataset are available in our Github repository along with the results of Mechanical Turk surveys.
\subsection{Details about Human Studies.}
\label{sec:appx-human}

\CAMTODO{Peruse and improve appendix.}

Our experiments relied on labor crowd-sourced from Amazon Mechanical Turk. We used five datasets: MIT and Yelp datasets from \cite{alzantot2018generating} and MIT, Yelp, and Movie Review datasets from \cite{BERT19}. We limited our worker pool to workers in the United States, Canada, Canada, and Australia that had completed over 5,000 HITs with over a 99\% success rate. We had an additional Qualification that prevented workers who had submitted too many labels in previous tasks from fulfilling too many of our HITs. In the future, will also use a small qualifier task to select workers who are good at the task.

For the human portions, we randomly select $100$ successful examples for each combination of attack method and dataset, then use Amazon's Mechanical Turk to gather $10$ answers for each example. For the automatic portions of the case study in Section \ref{sec:eval-methods}, we use all successfully perturbed examples. 

\subsubsection{Evaluating Adversarial Examples}

\paragraph{Rating Semantic Similarity.}\ In one task, we present results from two Mechanical Turk questionnaires to judge semantic similarity or dissimilarity. For each task, we show $x$ and $x_{adv}$, side by side, in a random order. We added a custom bit of Javascript to highlight character differences between the two sequences. We provided the following description: \say{Compare two short pieces of English text and determine if they mean different things or the same.} We then prompted labelers: \say{The changes between these two passages preserve the original meaning.} We paid $\$0.06$ per label for this task.

\paragraph{Inter-Annotator Agreement.}\ For each semantic similarity prompt, we gathered annotations from 10 different judges. Recall that each selection was one of 5 different options ranging from ``Strongly Agree" to ``Strongly Disagree." For each pair of original and perturbed sequences, we calculated the number of judges who chose the most frequent option. For example, if 7 choose ``Strongly Agree" and 3 chose ``Agree," the number of judges who chose the most frequent option is 7. We found that for the examples studied in Section \ref{sec:eval-methods} the average of this metric was $5.09$. For the examples in Section \ref{s5:better:constraints} at the threshold of $.98$ which we chose, the average was $5.6$.

\paragraph{Guessing Real vs. Computer-altered.}\ We present results from our Mechanical Turk survey where we asked users ``Is this text real or computer-altered?". We restricted this task to a single dataset, Movie Review. We chose Movie Review because it had an average sample length of 20 words, much shorter than Yelp or IMDB. We made this restriction because of the time-consuming nature of classifying long samples as Real or Fake. We paid $\$0.05$ per label for this task.

\paragraph{Rating word similarity.}\ We performed a third study where we asked showed users a pair of words and asked "In general, replacing the first word with the second preserves the meaning of a sentence:``. We paid $\$0.02$ per label for this task.

\paragraph{Phrasing matters.}\ Mechanical Turk comes with a set of pre-designed questionnaire interfaces. These include one titled ``Semantic Similarity'' which asks users to rate a pair of sentences on a scale from ``Not Similar At All'' to ``Highly Similar.'' Examples generated by synonym attacks benefit from this question formulation because humans tend to rate two sentences that share many words as ``Similar'' due to their small morphological distance, even if they have different meanings.

\paragraph{Notes for future surveys}.\ In the future, we would also try to filter out bad labels by mixing some number of ground-truth ``easy'' data points into our dataset and rejecting the work of labelers who performed poorly on this set.

\begin{figure*}[th]
\centering
\begin{minipage}[t]{.44\linewidth}
\includegraphics[width=\linewidth]{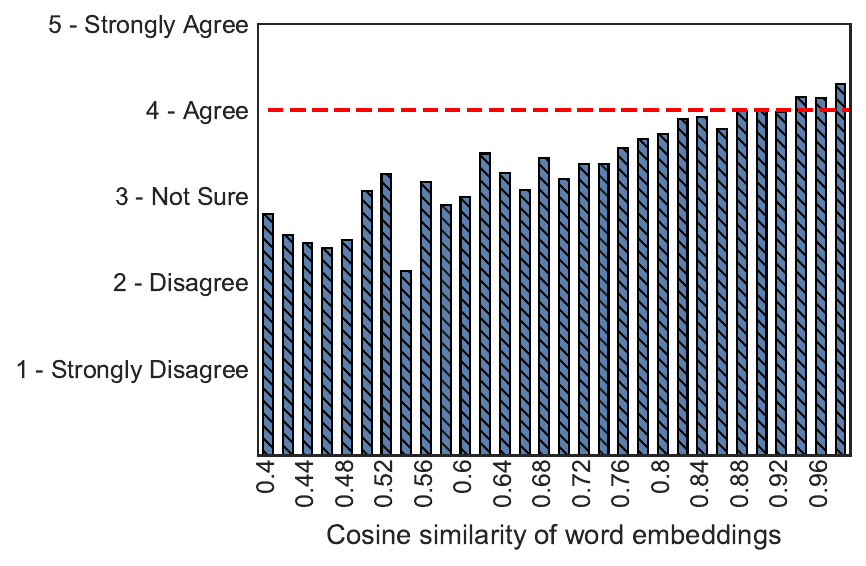}
\caption{Average response to ``In general, replacing the first word with the second preserves the meaning of a sentence'' vs. cosine similarity between word1 and word2 (words are grouped by cosine similarity into bins of size $.02$). \label{fig:s5-word-level-scores}}
\end{minipage}
\hfill
\begin{minipage}[t]{.48\linewidth}
\includegraphics[width=\linewidth]{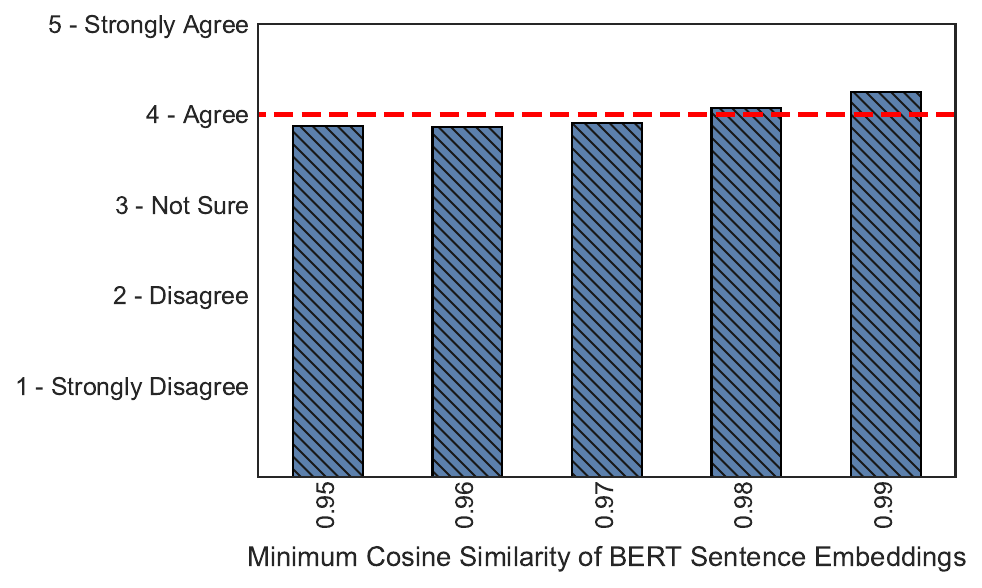}
\caption{Average response to ``The changes between these two passages preserve the original meaning" at each threshold. Threshold is minimum cosine similarity between BERT sentence embeddings. \label{fig:s5-sentence-level-scores}}
\end{minipage}
\end{figure*}

\subsubsection{Finding The Right Thresholds}
\label{sec:appx-thresholds}

\noindent \textbf{Comparing two words.} We showed study participants a pair of words and asked them whether swapping out one word for the other would change the meaning of a sentence. The results are shown in Figure \ref{fig:s5-word-level-scores}. Using this information, we chose $\textbf{0.9}$ as the word-level cosine similarity threshold.

\noindent \textbf{Comparing two passages.} With the word-level threshold set at $0.9$, we generated examples at sentence encoder thresholds 
$\{0.95, 0.96, 0.97, 0.98, 0.99\}$.  We chose to encode sentences with a pre-trained BERT sentence encoder fine-tuned for semantic similarity: first on the AllNLI dataset, then on the STS benchmark training set \cite{reimers-2019-sentence-bert}. We repeated the study from  \ref{sec:eval_semantics_human_s4}  on 100 examples from each threshold, obtaining 10 human labels per example. The results are in Figure  \ref{fig:s5-sentence-level-scores}. On average, judges agreed that the examples produced at $\textbf{0.98}$ threshold preserved semantics.

\subsection{Further Analysis of Non-Suspicious Constraint Case Study}
\label{appx:non-suspicious}

Table \ref{tab:conf_mat} presents the confusion matrix of results from the survey. Interestingly, workers guessed that the examples were real $62.2\%$ of the time, but when they guessed that examples were computer-altered they were right $75.4\%$ of the time. Thus while some perturbed examples are non-suspicious, there are some which workers identify with high precision. 

\begin{table}[t]
    \centering
    \begin{tabular}{cc|c|c|}
         \multicolumn{2}{c}{} & \multicolumn{2}{c}{Guessed Label}\\
         \multicolumn{2}{c}{} & \multicolumn{1}{c}{Real} & \multicolumn{1}{c}{Computer-altered} \\
         \cline{3-4}
         \multirow{2}{*}{True} & Original & 814 & 186 \\
         \cline{3-4}
          & Perturbed & 430 & 570 \\
          \cline{3-4}
    \end{tabular}
    \caption{Confusion matrix for humans guessing if perturbed examples are computer-altered}
    \label{tab:conf_mat}
\end{table}
\subsection{Adversarial Training Robustness Results}

\begin{figure*}[ht]
    \centering
    \includegraphics[width=\textwidth]{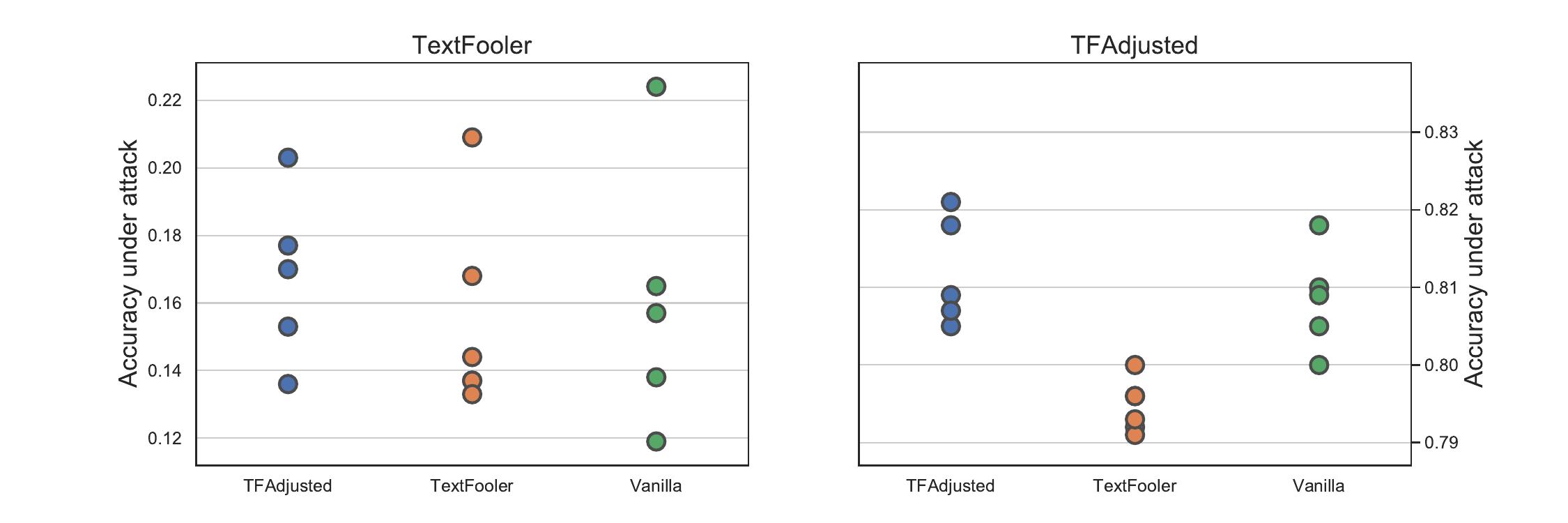}
    \caption{After-attack accuracy of our 15 adversarially trained models subject to two different attacks on the MR test set.}
    \label{fig:adv_retraining_robust}
\end{figure*}

We used our examples for adversarially training by attacking the full MR training set and retraining a new model with the successful examples appended to the training set. Previously, \citet{BERT19} reported an increase in robustness from adversarial training, while \cite{alzantot2018generating} reported no effect. We trained 5 models on each dataset, and saw significant variance in the robustness of adversarially trained models between random initializations and between epochs. The results are shown in Figure \ref{fig:adv_retraining_robust}. It is possible that \citet{BERT19} trained a single model for each training set (original and augmented) and happened to see an increase in robustness. It remains to be seen whether examples generated by \AlzName{}, \TfName{}, and \TfAdjName{} help or hurt the robustness and accuracy of adversarially trained models across other model architectures and datasets.

\CAMTODO{Repeat this experiment but with $>$ 10 models trained for $>$ 10 epochs, and on more datasets. We need more data.}
\subsection{Word Embeddings}

It is common to perform synonym substitution by replacing a word by a neighbor in the counter-fitted embedding space. The distance between word embeddings is frequently measured using Euclidean distance, but it is also common to compare word embeddings based on their cosine similarity (the cosine of the angle between them). (Some work also measures distance based on the mean-squared error between embeddings, which is just the square of Euclidean distance.)

For this reason, past work has sometimes constrained nearest neighbors based on the Euclidean distance between two word vectors, and other times based on their cosine similarity. \citet{alzantot2018generating} considered both distance metrics, and reported that they "did not see a noticeable improvement using cosine." 

We would like to point out that, when using normalized word vectors (as is typical for counter-fitted embeddings), filtering nearest neighbors based on their minimum cosine similarity \textbf{is equivalent to filtering by maximum Euclidean distance} (or MSE, for that matter). 

\begin{proof}
Let $u$, $v$ be normalized word embedding vectors. That is, $\lVert u\rVert=\lVert v\rVert=1$. Then $u \cdot v = \lVert u\rVert \lVert v\rVert \cos(\theta) = \cos(\theta)$. \\

{\centering
$  \displaystyle
    \begin{aligned}
    \lVert u-v\rVert ^2 &= (u - v) \cdot (u - v) \\
    &= \lVert u\rVert^2 - 2(u \cdot v) + \lVert v\rVert^2 \\
    &= 2 - 2(u \cdot v)\\
    &= 2 - 2\cos(\theta).
\end{aligned} $
\par}
\end{proof}

Therefore, the Euclidean distance between $u$ and $v$ is directly proportional to the cosine between them. For any minimum cosine distance $\eps$, we can use maximum euclidean distance $\sqrt{2 - 2\eps}$ and achieve the same result.
\\
\subsection{Examples In The Wild}

We randomly select 10 attempted attacks from the MR dataset and show the original inputs, perturbations before constraint change, and perturbations after constraint change. See Table \ref{tab:mr-examples}.

\begin{table*}[ht]
    \centering
 \begin{tabular}{|p{0.4\textwidth}|p{0.4\textwidth}|p{0.19\textwidth}|}
 \toprule
     \textbf{Original} & \textbf{Perturbed} (\TfName{}) & \textbf{Perturbed}  \\
      &  &  (\TfAdjName{}) \\
        \midrule
         by presenting an impossible romance in an impossible world , pumpkin dares us to say why either is impossible -- which forces us to confront what's possible and what we might do to make it so. {\color{green}Pos}: \textbf{99.5\%} & by presenting an {\color{purple}unsuitable} {\color{purple}romantic} in an impossible world , pumpkin dares {\color{purple}we} to say why either is {\color{purple}conceivable} -- which {\color{purple}vigour} {\color{purple}we} to {\color{purple}confronted} what's possible and what we might do to make it so. {\color{red}Neg}: \textbf{54.8\%} & {\color{gray}[Attack Failed]} \\ \hline
         
         ...a ho-hum affair , always watchable yet hardly memorable. {\color{red}Neg}: \textbf{83.9\%} & ...a ho-hum affair , always watchable yet {\color{purple}just} memorable. {\color{green}Pos}: \textbf{99.8\%} & {\color{gray}[Attack Failed]} \\ \hline
         
         schnitzler's film has a great hook , some clever bits and well-drawn, if standard issue, characters, but is still only partly satisfying. {\color{red}Neg}: \textbf{60.8\%}
         & schnitzler's film has a great hook, some clever {\color{purple}smithereens} and well-drawn, if standard issue, characters, but is still only partly satisfying. {\color{green}Pos}: \textbf{50.4\%} & schnitzler's film has a great hook, some clever {\color{cyan}traits} and well-drawn, if standard issue, characters, but is still only partly satisfying. {\color{green}Pos}: \textbf{56.9\%} 
         \\ \hline
         
         its direction, its script, and weaver's performance as a vaguely discontented woman of substance make for a mildly entertaining 77 minutes, if that's what you're in the mood for. {\color{green}Pos}: \textbf{99.5\%} & its direction, its script, and weaver's performance as a vaguely discontented woman of substance {\color{purple}pose} for a {\color{purple}marginally} {\color{purple}comical} 77 minutes, if that's what you're in the mood for. {\color{red}Neg}: \textbf{65.5\%} & {\color{gray}[Attack Failed]} \\ \hline
         
         missteps take what was otherwise a fascinating, riveting story and send it down the path of the mundane. {\color{green}Pos}: \textbf{99.1\%} & missteps take what was otherwise a fascinating, {\color{purple}scintillating} story and {\color{purple}dispatched} it down the path of the mundane. {\color{red}Neg}: \textbf{51.2\%} & {\color{gray}[Attack Failed]} \\ \hline
         
         hawke draws out the best from his large cast in beautifully articulated portrayals that are subtle and so expressive they can sustain the poetic flights in burdette's dialogue. {\color{green}Pos}: \textbf{99.9\%} & hawke draws out the {\color{purple}better} from his {\color{purple}wholesale} cast in {\color{purple}terribly} {\color{purple}jointed} portrayals that are {\color{purple}inconspicuous} and so expressive they can sustain the {\color{purple}rhymed} {\color{purple}flight} in burdette's dialogue. {\color{red}Neg}: \textbf{60.3\%} & {\color{gray}[Attack Failed]} \\ \hline
         
         if religious films aren't your bailiwick, stay away. otherwise, this could be a passable date film. {\color{red}Neg}: \textbf{99.1\%} & if religious films aren't your bailiwick, stay away. otherwise, this could be a {\color{purple}presentable} date film. {\color{green}Pos}: \textbf{86.6\%} & {\color{gray}[Attack Failed]} \\ \hline
         
         [broomfield] uncovers a story powerful enough to leave the screen sizzling with intrigue. {\color{green}Pos}: \textbf{99.1\%} & [broomfield] uncovers a story {\color{purple}pompous} enough to leave the screen sizzling with {\color{purple}plots}. {\color{red}Neg}: \textbf{59.2\%} & {\color{gray}[Attack Failed]} \\ \hline
         
         like its two predecessors, 1983's koyaanisqatsi and 1988's powaqqatsi, the cinematic collage naqoyqatsi could be the most navel-gazing film ever. {\color{green}Pos}: \textbf{99.4\%} & {\color{gray}[Attack Failed]} & {\color{gray}[Attack Failed]} \\ \hline
         
         maud and roland's search for an unknowable past makes for a haunting literary detective story, but labute pulls off a neater trick in possession : he makes language sexy. {\color{green}Pos}: \textbf{99.4\%} & maud and roland's search for an unknowable past makes for a haunting literary detective story, but labute pulls off a neater trick in {\color{purple}property} : he {\color{purple}assumes} language {\color{purple}sultry}. {\color{red}Neg}: \textbf{62.1\%} & {\color{gray}[Attack Failed]} \\
        \bottomrule
    \end{tabular}
    \caption{ Ten random attempted attacks, attacking BERT fine-tuned for sentiment classification on the MR dataset. Left column are original samples. Middle are perturbations with the constraint settings from \citet{BERT19}. Right column are perturbations generated with constraints adjusted to match human judgement. ``{\color{gray}[Attack Failed]}" denotes the {\color{gray}[Attack Failed]} to find a successful perturbation which fulfilled constraints. For 8 out of the 10 examples, the constraint adjustments caused the attack to fail.     \label{tab:mr-examples} }
\end{table*}

\end{document}